\documentclass[twocolumn]{article}
\usepackage{multirow,graphicx,blindtext,hyperref,booktabs,amsmath}
\hypersetup{
  colorlinks   = true, %Colours links instead of ugly boxes
  urlcolor     = blue, %Colour for external hyperlinks
  linkcolor    = blue, %Colour of internal links
  citecolor   = red %Colour of citations
}
\begin{document}

\title{Synthesising Multi-Modal Minority Samples for Tabular Data}
\author{Sajad Darabi\\
sajad.darabi@cs.ucla.edu\\
Amazon, New York
\and
Yotam Elor\\
yotam.elor@gmail.com\\
Amazon, New York}
\date{}

\maketitle
\begin{abstract}
Real-world binary classification tasks are in many cases imbalanced, where the minority class is much smaller than the majority class. This skewness is challenging for machine learning algorithms as they tend to focus on the majority and greatly misclassify the minority. Adding synthetic minority samples to the dataset before training the model is a popular technique to address this difficulty and is commonly achieved by interpolating minority samples. Tabular datasets are often multi-modal and contain discrete (categorical) features in addition to continuous ones which makes interpolation of samples non-trivial. To address this, we propose a latent space interpolation framework which (1) maps the multi-modal samples to a dense continuous latent space using an autoencoder; (2) applies oversampling by interpolation in the latent space; and (3) maps the synthetic samples back to the original feature space. We defined metrics to directly evaluate the quality of the minority data generated and showed that our framework generates better synthetic data than the existing methods. Furthermore, the superior synthetic data yields better prediction quality in downstream binary classification tasks, as was demonstrated in extensive experiments with 27 publicly available real-world datasets\footnote{Implementations of the oversamplers proposed here can be found at \href{https://github.com/aws/sagemaker-scikit-learn-extension/tree/master/src/sagemaker_sklearn_extension/contrib/taei}{https://github.com/aws/sagemaker-scikit-learn-extension/tree/master/src/sagemaker\_sklearn\_extension/cont-rib/taei}}.
\end{abstract}

\section{Introduction}
\label{sec:introduction}

Imbalanced classification tasks arise naturally, for example, consider the problem of credit card fraud detection where the vast majority of transactions are legitimate and only a few are fraudulent. This imbalance is challenging for machine learning (ML) algorithms as they tend to classify the majority well while poorly classifying the minority. This phenomenon occurs because the ML algorithms optimize a different metric than the user is interested in, resulting in an undesirable bias in the final trained model. Oversampling the minority class and under-sampling the majority class before training the ML algorithm are popular methods to address this challenge. They are effective because they yield an augmented dataset for which the algorithm's loss function and the desired loss function are more similar. See the formal description in Section \ref{sec:oversampling_overview}.

As opposed to random oversampling or increasing the weight of the minority class, \emph{SMOTE} was the first method to propose balancing the dataset by adding synthetic minority samples \cite{SMOTE}. In \emph{SMOTE}, the synthetic minority samples are created by interpolating pairs of the original minority points, hence instead of working in the original sample space by replicating samples, it generates new samples in the feature space. However, while effective for densely sampled feature spaces, When the feature space is sparse, the linear interpolation of samples might yield unrealistic low probability samples. Thus, \emph{SMOTE} only interpolates pairs of points that are relatively close in the feature space. However, this strategy is inefficient when the feature space is high-dimensional, see \cite{Lusa}.

Many real-life tabular datasets are multi-modal and include not only continuous numeric features but also categorical features e.g. gender, color, marital status, etc. We will denote the former continuous and the latter discrete. It is common to assume that the classification of each feature (continuous or discrete) is known. When interpolating samples of a multi-modal dataset with discrete features we face two challenges: (i) how to calculate distances between samples? and (ii) how to set the discrete feature values for the synthetic samples? In other words, how can we interpolate "dog" and "cat"? The original \emph{SMOTE} paper introduced \emph{SMOTE-NC} which is a \emph{SMOTE} variant that supports discrete features by: (i) using a heuristic to estimate the distance implied by the discrete features and (ii) the discrete feature values are set to be the majority of the $k$ nearest neighbors.

Since its' inception, more than $100$ extensions and variants of \emph{SMOTE} were proposed. However, to the best of our knowledge, \emph{SMOTE-NC} is the only variant that supports discrete features. A common method used to apply any interpolation method to discrete features is to encode them using ordinal integers and consider them to be continuous, see e.g. \cite{KOVACS2019105662}. This method results in synthetic samples that are continuous rather than discrete, thus not realistic. Moreover, many algorithms are optimized for handling discrete features (e.g. Catboost\cite{prokhorenkova2018catboost}, lightGBM\cite{lgbm} and deep networks\cite{arik2020tabnet}) and such augmentation will render these optimizations useless.

\begin{figure}
    \centering
    \includegraphics[width=\columnwidth]{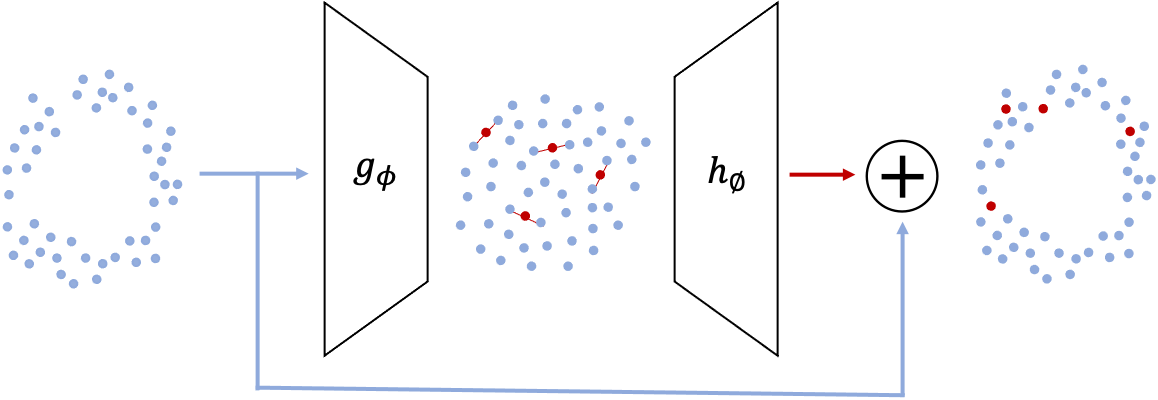}
    \caption{Overview of our proposed method. The sparse data is encoded into a dense latent space where synthetic samples (in red) are generated by interpolation. The interpolated samples are then decoded and added to the original data.}
    \label{fig:overview}
\end{figure}

Considering high-dimensional multi-modal data, it is commonly assumed that the data resides on an unknown lower dimensional manifold. For such sparse high-dimensional data, simple linear interpolation of samples can result in low probability synthetic samples. Motivated by this, we propose a latent space interpolation framework based on autoencoders that we name \emph{Tabular AutoEncoder Interpolator} (\emph{TAEI}). In summary, our method consists of a dimensionality reduction step where samples are mapped to a dense continuous latent space. Subsequently, samples of interest are interpolated in the learned latent space (using \emph{SMOTE} or any other interpolation technique) before being mapped back to the original feature space. Since our framework interpolates points in the dense latent space, it creates more genuine synthetic samples thus improving prediction performance. Additionally, our approach shifts the challenge of the discrete features from the interpolation method to the encoder-decoder, where we can leverage previous research. Thus, by mapping the discrete and continuous features to a unified continuous latent space, we enable dozens of previously proposed smote variants and extensions to produce multi-modal data. An overview of our method is shown in Figure \ref{fig:overview}.

When adding minority synthetic samples to the training data in order to improve prediction quality, intuitively, it's desired that the distribution of synthetic minority samples will mimic the underlying minority samples distribution. That is: (i) synthetic points will be generated wherever the underlying minority distribution is high and (ii) will not be generated where the underlying minority distribution is low. There is a trade-off between the two requirements above: For example, an oversampler can "play it safe" and create minority samples only where it has high confidence thus satisfying (ii) but not (i). Or, an oversampler can "take a risk" and create minority samples in larger regions thus satisfying (i) but not (ii). Two metrics are defined in Section \ref{sec:syntethic_evaluation} to capture these notions: \emph{cover} and \emph{error} for (i) and (ii) respectively. Then, by extensive experimentation with artificial data we map the methods proposed in the literature and our framework to the \emph{cover}-\emph{error} plane. As the best \emph{cover}-\emph{error} balance is unknown, we
identify the \emph{cover}-\emph{error} Pareto-optimal curve and show that the oversamplers of our framework constitutes its mid-range so other methods yield either much higher \emph{error}, worse \emph{cover} or sub-par performance for both. Furthermore, by experimentation with $27$ real-world binary classification datasets we show that the superior quality data generated by our proposed method yields better prediction quality for the downstream tasks.

To summarize, the contributions of our work are:
\begin{enumerate}
    \item We propose a new minority oversampling framework for tabular multi-modal data which outperforms state of the art oversampling techniques
    \item We introduce novel metrics to directly measure the quality of the generated data and demonstrate the effectiveness of our approach in generating high-quality synthetic data.
    \item Our framework is decomposable with any auto-encoding technique for multi-modal data and any continuous interpolation method, allowing it to take advantage of future advancements in both fields (auto-encoding and interpolation)
\end{enumerate}

\section{Related Work}
\label{sec:relatedwork}

Since \emph{SMOTE}'s inception, more than $100$ extensions and variations have been published. Due to their large number, we will not survey all \emph{SMOTE} variants but only the methods that resemble our approach. Moreover, to the best of our knowledge only \emph{SMOTE-NC} proposed in the original paper supports discrete features. In fact, the two recent survey papers do not even mention discrete features or multi-modal data, see \cite{SMOTE_SURVEY,KOVACS2019105662}. \cite{KOVACS2019105662} empirically compared $85$ variants of \emph{SMOTE} on $104$ imbalanced datasets including both continuous and discrete features. The paper does not describe how the discrete features were processed. We believe that they were simply treated as continuous after being encoded as ordinal integers. The best method was found to be \emph{Poly}\cite{Gazzah} which, similarly to \emph{SMOTE}, interpolates points in the feature space, but using a slightly different scheme. Differently from \emph{SMOTE}, \emph{Poly} allows interpolating of minority points that are not very close to each other. The second best performing algorithm, \emph{ProWSyn}\cite{Sukarna}, also allowed interpolating of far apart minority samples.

Several variants of \emph{SMOTE} share the idea of mapping the samples to another space which has some desired properties, interpolate in the new space and map the synthetic samples back to the original feature space. \cite{WANG4129201} proposed to map the samples using local linear embedding aiming to create a lower-dimensional space where the data  is  more  separable. \cite{Qiong} similarly proposed to use isometric feature  mapping (Isomap). Kernel functions were also used to map the features\cite{7222438,KernelADASYN}. When the classifier is an SVM, oversampling could be done directly in the kernel space\cite{KSMOTE}. 

As they are very natural, autoencoders were previously proposed as a means to provide the bi-directional mapping. \cite{Bellinger} proposed to create synthetic samples by adding Gaussian noise to the original samples in the latent space and then decode them back to the feature space. Later, the same authors proposed to apply \emph{SMOTE} in the latent space, see \cite{Bellinger2016BeyondTB}. However, there are two key differences between their approach and ours: (1) they train the autoencoder on the minority class samples only which, due to the low number of samples, force them to train shallow models and (2) they did not consider discrete features. \cite{Babaei2019DataAB} proposed to encode the samples and train the classifier in the latent space. To improve unsupervised anomaly detection, \cite{doping} proposed to use adversarial auto encoders to encode the features into a Gaussian mixture latent space and interpolate samples near the boundaries of the distribution. Some methods incorporate mapping the feature to other spaces but use the new space differently. MOT2LD\cite{Zhipeng} first maps each training sample into a low-dimensional space and then applies clustering and weighting heuristics in the low-dimensional space. In ADOMS\cite{TANG4570642} each sample neighbors are derived in the original feature space, however, the synthetic sample is then created along with the first principal component of the k neighbors.

Synthetic data generation has been studied extensively in the context of deep learning for image collections. For example, \cite{Zhang2018mixupBE} is an interpolating framework very similar to SMOTE intended for deep networks. Interpolation in the latent space using autoencoders was also proposed~\cite{yin2019feature,carter2017using}. Similarly to our work, the quality of the resulting synthetic samples was directly studied in \cite{berthelot2018understanding} using artificial data. However, the study on image collection is not easily transferable to tabular datasets as image collections have different characteristics. For example, image data is spatially coherent. Furthermore, image collections usually contain hundreds of thousands of images while the number of samples in our tabular datasets is usually two or three orders of magnitude smaller. In fact, the median and average number of samples in our dataset collection are about $10k$ and $3k$ respectively.

Nevertheless, inspired by study of image collections, generation of tabular multi-modal data was studied in the context of generative adversarial networks (GANs). The challenge of synthesizing the discrete features was addressed using three methods: noising the discrete data~\cite{xu2018synthesizing}, using a Gumbel softmax~\cite{NIPS2019_8953,CTGAN} or using an autoencoder to map the data to a continuous latent space and train the GAN in that space~\cite{pmlr-v68-choi17a}. FAST-DAD~\cite{fakoor2020fast} generates multi-modal synthetic samples by augmenting existing samples using Gibbs sampling. The Gibbs based augmentation method requires a pre-trained conditional expectation model for all features and another model is used to label the new samples.

\section{Method}
\label{sec:method}
\subsection{Problem Statement}
\label{sec:oversampling_overview}

Consider a dataset $(X,Y) = \{(x_i, y_i)\}_{i=1}^N$, where $x_i$ are observations sampled from a data-generating distribution $p(x)$ and $y_i \in \{0, 1\}$. We consider a multi-modal setting where $x_i$ is a concatenation of discrete $\mathcal{D} = [D_1, \cdots, D_{|\mathcal{D}|}]$ and continuous features $\mathcal{C} = [C_1, \cdots, C_{|\mathcal{C}|}]$. In classification tasks, the goal is to obtain a classifier function $f : \mathcal{X} \rightarrow \mathcal{Y} \in \mathcal{F}$ optimizing a given loss function $l(\cdot,\cdot)$
\begin{equation*}
    f = \min_{f\in \mathcal{F}}l\left(f(X),Y\right)
\end{equation*}
Where $f(X)$ is the vector of predictions or class probabilities and $Y$ is the vector of true labels. For imbalanced binary classification, $l(\cdot,\cdot)$ is commonly ROC-AUC, $F_1$-score or $G$-score.

\subsection{Method Overview}

Our proposed framework is based on the standard autoencoder scheme proposed in \cite{hinton2006reducing}. As we are concerned with multi-modal data, for every categorical column $D$, we introduce an embedding layer $\mathcal{W} \in \Re^{|D|\times d_{|D|}}$. The input which is a concatenation of discrete and continuous features is passed through the embedding layer resulting in a feature vector $x_i \in \Re^{|\mathcal{C}| + \sum_i^{|D|}d_{|D_i|}}$ that is used as the input to the autoencoder
\begin{equation*}
    z = g_\phi(x)
\end{equation*}
Where $g_\phi(\cdot)$ is a high capacity deep neural network, such as a set of fully connected layers.

As in traditional autoencoders, a decoder module $h_\theta(z)$ is used to map the latent points back to the feature space $x$. The objective minimized while training the autoencoder is the reconstruction loss:
\begin{equation*}
   \min_{\theta,\phi}{E_{p(x)}[d(h_\theta(z), x)]}
\end{equation*}
In a multi-modal setting, the reconstruction contains both discrete and continuous variables therefore the function $d(\cdot)$ is a sum of the softmax loss and mean squared error (MSE):
\begin{align*}
    J_{recon}(D;\theta, \phi) &= \sum_{x_i} \sum_c^{|C|} ||h_\theta(z_i)^c - x_i^c||_2^2\\
    & + \alpha \sum_{x_i}{\sum_d^{|D|}\sum_o{ \mathbf{1}[x_i^d = o]\log(h_\theta(z_i)^o)}}
\end{align*}
Where $\alpha$ is a constant controlling the weight of the softmax loss. In our experiments $\alpha=0.3$ was used.

Once we have a fully trained autoencoder we leverage existing interpolation methods to synthesize minority samples in the latent space and using the decoder to map them back to the feature space. In our experiments \emph{SMOTE} and \emph{Poly}\cite{Gazzah} were used. \emph{Poly} is a simple interpolation technique that was selected based on it's superior performance in the experiments of \cite{KOVACS2019105662}. Nevertheless, our framework and the published code support any underlying interpolation method.

\subsection{Autoencoder Schemes}
\label{sec:flavors}

As samples are interpolated in the latent space, it is desired that the latent space manifold will be dense and without holes i.e. without low probability regions surrounded by high probability ones. When the latent space manifold contains holes, pairs of points from high probability regions might by interpolated into these low probability regions which can result in the generation of unrealistic samples. Vanilla autoencoders do not have such guarantees, hence we studied several other autoencoders with the motivation of creating a dense latent space in which linear interpolation of samples from the manifold will reside on the manifold as well.

\vspace{3pt}
\noindent\textbf{Variational autoencoder (VAE)} Due to their probabilistic nature, VAEs naturally yield a smooth latent space manifold. We have used the vanilla VAE\cite{kingma2014autoencoding} with embeddings for the discrete columns and $J_{recon}$ as defined above.

\vspace{3pt}
\noindent\textbf{Regularized autoencoder (RAE)} Recent developments have shown that adding a regularization loss in the latent space results in a deterministic autoencoder with similar properties as in VAEs \cite{rae}, all be it much faster and easier to train. Thus, in RAE, norm $2$ regularization is added to the latent space representation ($z$) and to the decoder weights.

\vspace{3pt}
\noindent\textbf{Adversarial autoencoder (AAE)} Similarly to \cite{doping}, we train the autoencoder in an adversarial setting vs a discriminator in the latent space. The discriminator tries to classify real samples latent representations vs vectors drawn randomly from a Gaussian distribution and the autoencoder aims at minimizing $J_{recon}$ while fooling the discriminator.

\vspace{3pt}
\noindent\textbf{Interpolate adversarial autoencoder (IAAE)} IAAE is introduced in order to directly penalize invalid interpolation in the latent space. It is achieved in an adversarial setting using a discriminator with the task of classifying real samples latent space representations vs latent space interpolated point. In order to fool the discriminator, the encoder learns a latent representation in which interpolated samples are similar to real ones

\vspace{3pt}
\noindent\textbf{Adversarially constrained autoencoder interpolation (ACAI)} Adaptation of the autoencoder proposed in \cite{berthelot2018understanding} for tabular datasets. The autoencoder is regularized by mixing two samples with ratio $\alpha$ and $1-\alpha$ in the latent space. The adversarial critic network tries to predict $\alpha$ given only the decoded interpolated sample.

% \vspace{3pt}
% All the autoencoder schemes described above are unsupervised in the sense that they do not use the labels in training the autoencoder. The CAE oversampler described below use the labels in training to better separate the classes in the latent space.

% \vspace{3pt}
% \noindent\textbf{Contrastive loss autoencoder (CAE)} Contrastive loss has been shown recently to yield state of the art results for unsupervised and self-supervised learning\cite{chen2020simple,misra2019selfsupervised}. In our implementation, the contrastive loss is derived by selecting a single minority sample from the mini-batch and calculating it's average distance to all other minority samples (in the mini-batch) contrasted with the average distance to all majority samples. Formally, the loss is given by:
% \begin{equation*}
%     CL = - log\left( \frac
%     {\frac{1}{Z_T}\sum_{z\in Z_T}{\log\left(sim\left(f(z_0),f(z)\right) / \tau\right)}}
%     {\frac{1}{Z_F}\sum_{z\in Z_F}{\log\left(sim\left(f(z_0),f(z)\right) / \tau\right)}}
%     + \epsilon \right)
% \end{equation*}
% where $z_0$ is the latent representation of the selected minority sample, $Z_T$ and $Z_F$ are the remaining minority and majority samples in the mini-batch respectively, $f(\cdot)$ is a deep model, $sim(\cdot,\cdot)$ is the cosine distance, $\tau$ is the temperature constant and $\epsilon$ is a small constant added for numeric stability. See more details in Appendix \ref{details:CAE}.

\section{Evaluation on Artificial Data}
\label{sec:syntethic_evaluation}

In the spirit of \cite{berthelot2018understanding}, we use artificial datasets to evaluate the quality of the synthetic minority samples generated by the oversamplers. As opposed to real datasets, for artificial ones we know the underlying distribution thus we can directly measure the quality of the synthetic data. Generally, artificial datasets are created by defining the minority and majority distributions in feature space and sampling from them. In our experiments these distributions are defined by two non-overlapping manifolds: the \emph{minority manifold} and the \emph{majority manifold}. The artificial datasets are created by sampling uniformly from these two manifolds. To create multi-modal data, the artificial samples are partially discretized as described below.

To create artificial sparse data, we (1) define a low-dimension manifold in feature space and (2) split the manifold into \emph{minority manifold} and \emph{majority manifold}. So it is guaranteed that the minority and majority samples reside in separate regions of the same low-dimension manifold. For a $d$-dimensional feature space, the manifold is a $d$-dimensional unit sphere (which is a $(d-1)$-dimensional manifold). The \emph{minority manifold} is a thin slice of the sphere. Formally, for a $D$-dimensional feature space the \emph{minority manifold} is defined by $||X|| = 1\,\wedge\,|x_0|\leq\alpha$ and the \emph{majority manifold} is given by $||X|| = 1\,\wedge\,|x_0|>\alpha$ where $||\cdot||$ is the euclidean norm and $x_0$ is the first coordinate of the vector $X$. $\alpha$ is a constant set so the density of both manifolds will be the same. For example, when the minority samples are $5\%$ of the samples as was used in the experiments, $\alpha=0.06$.

\begin{table*}
    \footnotesize
    \begin{minipage}{.33\textwidth}
         \begin{tabular}{clll}
& & \textbf{\emph{Cov}} & \textbf{\emph{Err}}\\ \toprule 
\parbox[t]{0pt}{\multirow{5}{*}{\rotatebox[origin=c]{90}{\tiny Pareto-optimal}}}& TGAN & 0.323 & 0.617\\ 
& \textbf{Poly+VAE} & 0.335 & 0.301\\ 
& \textbf{SMOTE+VAE} & 0.355 & 0.138\\ 
& \textbf{SMOTE+AE} & 0.378 & 0.082\\ 
& No OS & 0.449 & 0.000\\ 
\midrule 
& CTGAN & 0.355 & 0.683\\ 
& \textbf{Poly+RAE} & 0.364 & 0.211\\ 
& \textbf{Poly+ACAI} & 0.367 & 0.225\\ 
& \textbf{Poly+AE} & 0.370 & 0.187\\ 
& \textbf{SMOTE+RAE} & 0.378 & 0.094\\ 
& SMOTE/NC & 0.380 & 0.121\\ 
& \textbf{SMOTE+ACAI} & 0.384 & 0.110\\ 
& \textbf{Poly+IAAE} & 0.388 & 0.321\\ 
& \textbf{Poly+AAE} & 0.395 & 0.393\\ 
& \textbf{SMOTE+AAE} & 0.400 & 0.315\\ 
& \textbf{SMOTE+IAAE} & 0.414 & 0.264\\ 
\\ 
\bottomrule
\end{tabular}
        \begin{center}(a) All\end{center}
    \end{minipage}
    \begin{minipage}{.33\textwidth}
         \begin{tabular}{clll}
& & \textbf{\emph{Cov}} & \textbf{\emph{Err}}\\ \toprule 
\parbox[t]{0pt}{\multirow{6}{*}{\rotatebox[origin=c]{90}{\scriptsize Pareto-optimal}}}& \textbf{Poly+VAE} & 0.337 & 0.187\\ 
& \textbf{SMOTE+VAE} & 0.364 & 0.102\\ 
& Poly & 0.370 & 0.063\\ 
& \textbf{SMOTE+AE} & 0.393 & 0.044\\ 
& SMOTE/NC & 0.405 & 0.025\\ 
& No OS & 0.463 & 0.000\\ 
\midrule 
& TGAN & 0.342 & 0.622\\ 
& CTGAN & 0.367 & 0.673\\ 
& \textbf{Poly+RAE} & 0.377 & 0.140\\ 
& \textbf{Poly+ACAI} & 0.381 & 0.166\\ 
& \textbf{Poly+AE} & 0.389 & 0.097\\ 
& \textbf{SMOTE+RAE} & 0.398 & 0.054\\ 
& \textbf{Poly+IAAE} & 0.407 & 0.292\\ 
& \textbf{SMOTE+ACAI} & 0.408 & 0.075\\ 
& \textbf{SMOTE+IAAE} & 0.421 & 0.221\\ 
& \textbf{Poly+AAE} & 0.430 & 0.362\\ 
& \textbf{SMOTE+AAE} & 0.433 & 0.302\\ 
\bottomrule
\end{tabular}
        \begin{center}(b) Continuous\end{center}
    \end{minipage}
    \begin{minipage}{.33\textwidth}
         \begin{tabular}{clll}
& & \textbf{\emph{Cov}} & \textbf{\emph{Err}}\\ \toprule 
\parbox[t]{0pt}{\multirow{6}{*}{\rotatebox[origin=c]{90}{\scriptsize Pareto-optimal}}}& TGAN & 0.303 & 0.612\\ 
& \textbf{Poly+VAE} & 0.333 & 0.414\\ 
& \textbf{SMOTE+VAE} & 0.347 & 0.174\\ 
& \textbf{SMOTE+RAE} & 0.357 & 0.134\\ 
& \textbf{SMOTE+AE} & 0.362 & 0.120\\ 
& No OS & 0.435 & 0.000\\ 
\midrule 
& CTGAN & 0.342 & 0.694\\ 
& \textbf{Poly+RAE} & 0.351 & 0.283\\ 
& \textbf{Poly+AE} & 0.352 & 0.277\\ 
& \textbf{Poly+ACAI} & 0.353 & 0.285\\ 
& SMOTE/NC & 0.355 & 0.217\\ 
& \textbf{SMOTE+ACAI} & 0.360 & 0.144\\ 
& \textbf{Poly+AAE} & 0.360 & 0.424\\ 
& \textbf{SMOTE+AAE} & 0.367 & 0.328\\ 
& \textbf{Poly+IAAE} & 0.369 & 0.351\\ 
& \textbf{SMOTE+IAAE} & 0.406 & 0.307\\ 
\\ 
\bottomrule
\end{tabular}
        \begin{center}(c) Multi-modal\end{center}
    \end{minipage}
\caption{\emph{Cover} and \emph{error} of the oversamplers averaged over artificial datasets. (a) results averaged over all artificial datasets; (b) results averaged over all continuous artificial datasets; (c) results averaged over all multi-modal artificial datasets. The oversamplers of the Pareto-optimal front are above the mid-line and our proposed methods are in bold.
}
    \label{tab:synth_pareto_optimal}
\end{table*}

To create a continuous artificial dataset of size $N$ with $\mu$ of the samples being minority: (1) sample uniformly $N(1-\mu)$ majority samples from the \emph{majority manifold} and $N\mu$ minority samples from the \emph{minority manifold}, and (2) rotate the sphere randomly. To create a multi-modal dataset with both continuous and discrete features: (1) create a continuous artificial dataset, (2) discretize some of the features using $M$ bins equally spaced between $-1$ and $1$, and (3) randomly change the order of bins. The order of bins is randomized in order to remove the geometric information from the ordinal representation i.e., avoid bin $0$ being the leftmost, $1$ being the one next to it and so on. We experimented with $D\in[6, 10]$, $N\in[1k, 10k, 100k]$, $M=7$ and $\mu=0.05$ so we had $6$ continuous and $6$ multi-modal classes of artificial datasets. In the multi-modal datasets half of the features were discretized. In our experiments, $7$ artificial datasets were generated for each of the $12$ classes using different random seeds and the results were averaged over them.

As previously discussed, when synthesising minority data for classification, it's desired that the distribution of synthetic minority samples will mimic the underlying minority samples distribution. That is: (i) synthetic points will be generated wherever the underlying minority distribution is high and (ii) will not be generated where the underlying minority distribution is low. However, generating synthetic minority samples in regions in feature space that have low minority density and high majority density is more harmful than generating them in regions with low minority and majority densities. Thus, we reformulate (ii) as: do not generate minority samples in regions where the majority density is higher than the minority density. The metrics used to measure these notions are formally defined as:

\vspace{3pt}
\noindent\textbf{\emph{Cover} (\emph{Cov})} measures how well the synthetic minority samples cover the \emph{minority manifold}. \emph{Cover} is calculated by (1) using the oversampler to create $10k$ synthetic minority samples, (2) sampling $500k$ minority samples from the \emph{minority manifold}, (3) for each "real" minority sample, calculate the distance to the closest synthetic point, (4) cover is defined as the average of these distances. Note that for \emph{cover}, lower is better.

\vspace{3pt}
\noindent\textbf{\emph{Error} (\emph{Err})} quantifies how much the synthetic minority samples are reliable as minority. It is calculated by (1) using the oversampler to create $10k$ synthetic minority samples, (2) sampling $500k$ minority samples from the \emph{minority manifold} and $500k$ majority samples from the \emph{majority manifold}, (3) for each synthetic sample find the closest "real" sample. A synthetic sample is \emph{invalid} if the closest "real" point is a majority point. \emph{Error} is defined as the ratio of invalid synthetic points to the number of synthetic points generated.
\vspace{3pt}

Both \emph{cover} and \emph{error} rely on a distance metric in the feature space. Euclidean distance was used for continuous feature spaces. For multi-modal, as they have an underlying geometry, the discrete features where converted back to the continuous space before calculating the Euclidean distance. It was achieved by replacing each discrete feature value with the center of the bin previously used for discretizing. It was done for both the generated synthetic samples and the "real" ones.

As previously discussed,  there is a trade-off between optimizing \emph{cover} and \emph{error}. An oversampler can "play it safe" and create minority samples only where it has high confidence thus yielding a lower \emph{error} but also a higher (worse) \emph{cover}. On the other hand, an oversampler can "take a risk" and create minority samples in larger regions thus lowering (improving) \emph{cover} but increasing the \emph{error}. Since the preffered balance is unclear, and the optimal \emph{cover}-\emph{error} ratio probably varies for different datasets, we are interested in the Pareto-optimal front.

Recall that there are very few methods that allow generation of multi-modal tabular data including both continuous and discrete features. We compared our autoencoder based oversamplers to all available methods:

\vspace{3pt}
\noindent\textbf{\emph{SMOTE/NC}}: As \emph{SMOTE} supports only continuous datasets and \emph{SMOTE-NC} supports only datasets that have both numeric and discrete features (fails otherwise) we paired them together under the name \emph{SMOTE/NC}.

\vspace{3pt}
\noindent\textbf{CTGAN}\cite{CTGAN}: A recent generative GAN model specifically designed to handle tabular datasets by conditioning on discrete columns. The open source implementation was used\footnote{\href{https://github.com/sdv-dev/CTGAN}{https://github.com/sdv-dev/CTGAN}}.

\vspace{3pt}
\noindent\textbf{TGAN}\cite{arjovsky2017wasserstein}: The TGAN network trained using Wassertein GAN loss. The open source implementation was used\footnote{\href{https://github.com/Baukebrenninkmeijer/On-the-Generation-and-Evaluation-of-Synthetic-Tabular-Data-using-GANs}{https://github.com/Baukebrenninkmeijer/On-the-Generation-and-Evaluation-of-Synthetic-Tabular-Data-using-GANs}}.

\vspace{3pt}
\noindent\textbf{\emph{Poly}}\cite{Gazzah}: An interpolation method that supports only continuous features thus was included only in the experiments with continuous artificial datasets. \emph{Poly} was included as it performed the best amongst the 85 \emph{SMOTE} variations evaluated in \cite{KOVACS2019105662}.

\vspace{3pt}
\noindent\textbf{\emph{No OS}}: Not oversampling - using the original minority points. In addition to providing a baseline, the random oversampling used in Section \ref{sec:real_data_evaluation} yields the same \emph{cover} and \emph{error} as \emph{No OS}.
\vspace{3pt}

As CTGAN and TGAN generate both majority and minority samples, they were used by: (1) considering the label to be an additional discrete feature in the training phase; (2) when oversampling, generate more samples then required and filter out the majority samples; (3) repeat the previous step until the desired number of minority samples is generated.

\begin{figure}
    \centering
    \includegraphics[width=0.9\columnwidth]{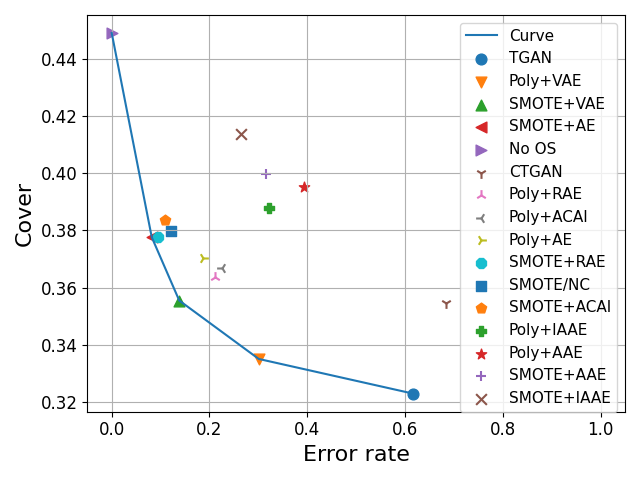}
    \caption{\emph{Cover} and \emph{error} for the oversamplers averaged over all artificial datasets. The Pareto-optimal curve is presented by a blue line.}
    \label{fig:par_opt}
\end{figure}

The resulting \emph{Cover} and \emph{error} of the oversamplers are presented in Table \ref{tab:synth_pareto_optimal}. The Pareto-optimal oversamplers are above the mid-line. The results averaged over all artificial datasets can be found in Table \ref{tab:synth_pareto_optimal}a and Figure \ref{fig:par_opt}. The results averaged over several slices of the artificial datasets can be found in Tables \ref{tab:synth_pareto_optimal}b, \ref{tab:synth_pareto_optimal}c and \ref{tab:synth_100k}.

Overall, TGAN yields the best \emph{cover}. However, this comes with the price of very high \emph{error}. \emph{No OS} yields the worst \emph{cover} but with an \emph{error} of zero, hence it's always a part of the Pareto-optimal front. From our proposed autoencoders, the VAE and AE based oversamplers consistently outperform the others. Overall (Table \ref{tab:synth_pareto_optimal}a) VAE and AE fill the mid-range between TGAN and \emph{No OS} and outperform all other oversamplers in that range. \emph{Poly} supports only continuous features hence it was omitted for multi-modal datasets. However, for continuous datasets it performs well and belongs to the Pareto-optimal front, see Table \ref{tab:synth_pareto_optimal}b.

We postulated that the GAN based methods, TGAN and CTGAN, would benefit from training on a large datasets. However, this was not the case as the results are consistent for large and small datasets, see Table \ref{tab:synth_100k} for the results on artificial datasets comprising $100k$ samples.

\begin{figure}
    \begin{center}
    \begin{minipage}{.49\columnwidth}
    \includegraphics[width=\textwidth]{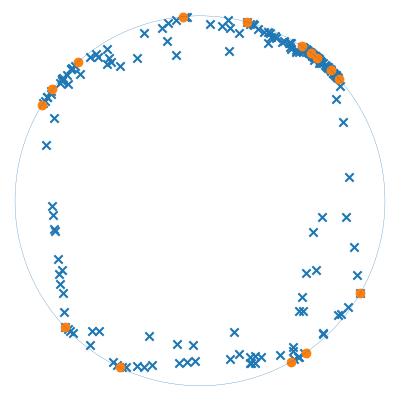}
    \begin{center}(a) \emph{SMOTE}\end{center}
    \end{minipage}
    \begin{minipage}{.49\columnwidth}
    \includegraphics[width=\textwidth]{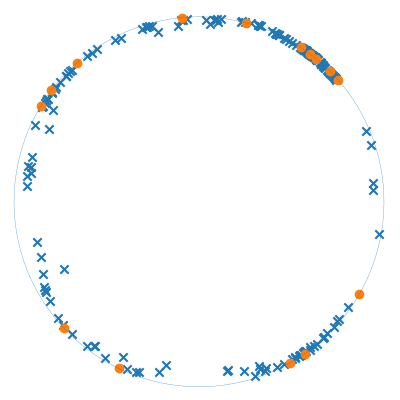}
    \begin{center}(b) Our framework\end{center}
    \end{minipage}
    \caption{An example of the synthetic minority samples generated by \emph{SMOTE} and our framework. The artificial manifold is a $3d$ sphere thus the minority manifold resembles a $2d$ ring. The real minority samples are marked by orange dots and the generated synthetic samples by blue $\times$s. See (a) where \emph{SMOTE} generates some low probability sample inside the ring. On the other hand, see (b), where our framework by interpolating samples in the dense latent space generates realistic samples i.e., closer to the underlying minority manifold, on the ring as opposed to in it.}
    \label{fig:example}
    \end{center}
\end{figure}

An example of our framework generating more realistic synthetic data compared to \emph{SMOTE} is presented in Figure \ref{fig:example}. 

\begin{table}
    \footnotesize
    \centering
    \begin{minipage}{.33\textwidth}
         \begin{tabular}{clll}
& & \textbf{\emph{Cov}} & \textbf{\emph{Err}}\\ \toprule 
\parbox[t]{0pt}{\multirow{4}{*}{\rotatebox[origin=c]{90}{\scriptsize P-optimal}}}& \textbf{SMOTE+VAE} & 0.265 & 0.084\\ 
& \textbf{SMOTE+RAE} & 0.274 & 0.043\\ 
& \textbf{SMOTE+AE} & 0.275 & 0.031\\ 
& No OS & 0.287 & 0.000\\ 
\midrule 
& SMOTE/NC & 0.276 & 0.061\\ 
& \textbf{Poly+VAE} & 0.283 & 0.326\\ 
& \textbf{SMOTE+ACAI} & 0.290 & 0.053\\ 
& \textbf{Poly+RAE} & 0.302 & 0.204\\ 
& \textbf{Poly+ACAI} & 0.306 & 0.162\\ 
& \textbf{SMOTE+AAE} & 0.311 & 0.260\\ 
& TGAN & 0.314 & 0.617\\ 
& \textbf{Poly+AE} & 0.323 & 0.153\\ 
& CTGAN & 0.339 & 0.645\\ 
& \textbf{Poly+AAE} & 0.360 & 0.357\\ 
& \textbf{SMOTE+IAAE} & 0.385 & 0.361\\ 
& \textbf{Poly+IAAE} & 0.391 & 0.386\\
\\
\bottomrule
\end{tabular}
    \end{minipage}
\caption{\emph{Cover} and \emph{error} of the oversamplers averaged over artificial datasets with $100k$ samples. The oversamplers of the Pareto-optimal front are above the mid-line and our proposed methods are in bold.
}
    \label{tab:synth_100k}
\end{table}

\section{Evaluation on Real-World Data}
\label{sec:real_data_evaluation}

\begin{table*}
    \footnotesize

\begin{tabular}{lcccccccc} 
 & \textbf{abalone} & \textbf{abalone\_19} & \textbf{arrhythmia} & \textbf{car\_eval\_34} & \textbf{car\_eval\_4} & \textbf{coil\_2000} & \textbf{ecoli} & \textbf{isolet}\\ \toprule 
No OS & 0.8617 & 0.733 & 0.968 & 0.9973 & 0.9985 & 0.745 & 0.948143 & 0.9913\\ 
CTGAN & 0.8617 & 0.713 & 0.981 & 0.9978 & 0.9987 & 0.749 & \textbf{0.96554} & 0.9916\\ 
Poly & 0.8632 & 0.767 & \textbf{0.987} & 0.9985 & \textbf{0.9996} & 0.746 & 0.960856 & 0.9918\\ 
\textbf{Poly+AE} & 0.8621 & 0.769 & 0.975 & 0.9978 & 0.9989 & 0.753 & 0.959518 & \textbf{0.9924}\\ 
\textbf{Poly+VAE} & 0.8617 & \textbf{0.79} & 0.985 & \textbf{0.9988} & 0.9995 & \textbf{0.756} & 0.96554 & 0.9916\\ 
ROS & 0.8632 & 0.738 & 0.983 & 0.9979 & 0.9993 & 0.749 & 0.94898 & 0.9919\\ 
SMOTE & 0.8632 & 0.753 & 0.985 & 0.9984 & 0.9989 & 0.746 & 0.963031 & 0.9919\\ 
\textbf{SMOTE+AE} & \textbf{0.8635} & 0.773 & 0.985 & 0.9981 & 0.9992 & 0.752 & 0.960355 & 0.9918\\ 
\textbf{SMOTE+VAE} & 0.863 & 0.749 & 0.984 & 0.9981 & 0.9989 & 0.755 & 0.960522 & 0.9912\\ 
SMOTE-NC & 0.863 & 0.735 & 0.983 & \multicolumn{1}{c}{-} & \multicolumn{1}{c}{-} & 0.748 & \multicolumn{1}{c}{-} & 0.9921\\ 
TGAN & 0.8618 & 0.76 & 0.978 & 0.998 & 0.9985 & 0.748 & 0.962529 &\\ 
\end{tabular}

\begin{tabular}{lccccccc} 
 & \textbf{letter\_img} & \textbf{libras\_move} & \textbf{mammography} & \textbf{oil} & \textbf{optical\_digits} & \textbf{ozone\_level} & \textbf{pen\_digits}\\ \toprule 
No OS & 0.99956 & 0.981 & 0.9557 & 0.9 & 0.999 & 0.891 & 0.99978\\ 
CTGAN & 0.99958 & 0.987 & 0.9568 & 0.912 & 0.999 & 0.895 & 0.99983\\ 
Poly & 0.99963 & 0.988 & 0.9578 & 0.932 & 0.9992 & 0.909 & 0.99985\\ 
\textbf{Poly+AE} & \textbf{0.99975} & \textbf{0.994} & 0.9579 & 0.917 & 0.9992 & 0.905 & \textbf{0.99986}\\ 
\textbf{Poly+VAE} & 0.99962 & 0.992 & \textbf{0.9588} & 0.912 & 0.9992 & \textbf{0.915} & 0.99984\\ 
ROS & 0.9995 & 0.986 & 0.9576 & 0.929 & 0.9992 & 0.901 & 0.99982\\ 
SMOTE & 0.99955 & 0.99 & 0.956 & \textbf{0.944} & \textbf{0.9993} & 0.904 & 0.99982\\ 
\textbf{SMOTE+AE} & 0.99966 & 0.988 & 0.9587 & 0.938 & 0.9991 & 0.899 & 0.99983\\ 
\textbf{SMOTE+VAE} & 0.9996 & 0.992 & 0.9583 & 0.919 & 0.9991 & 0.908 & 0.99984\\ 
SMOTE-NC & \multicolumn{1}{c}{-} & \multicolumn{1}{c}{-} & \multicolumn{1}{c}{-} & 0.935 & 0.9992 & \multicolumn{1}{c}{-} & \multicolumn{1}{c}{-}\\ 
TGAN & 0.99957 & 0.985 & 0.9558 & 0.902 & 0.999 & 0.878 & 0.99981\\ 
\end{tabular}

\begin{tabular}{lcccccc} 
 & \textbf{protein\_homo} & \textbf{satimage} & \textbf{scene} & \textbf{sick\_euthyroid} & \textbf{solar\_flare\_m0} & \textbf{spectrometer}\\ \toprule 
No OS & 0.9922 & 0.955 & 0.766 & 0.975 & 0.819 & 0.959\\ 
CTGAN & 0.993 & 0.959 & 0.782 & 0.975 & 0.827 & 0.976\\ 
Poly & \textbf{0.9935} & 0.961 & 0.775 & 0.969 & 0.816 & 0.981\\ 
\textbf{Poly+AE} & 0.9931 & 0.96 & 0.783 & \textbf{0.976} & 0.817 & 0.973\\ 
\textbf{Poly+VAE} & 0.9922 & 0.962 & \textbf{0.799} & 0.975 & \textbf{0.829} & 0.971\\ 
ROS & 0.9925 & 0.961 & 0.782 & 0.973 & 0.818 & 0.959\\ 
SMOTE & 0.9925 & \textbf{0.963} & 0.778 & 0.972 & 0.811 & 0.97\\ 
\textbf{SMOTE+AE} & 0.9934 & 0.962 & 0.786 & 0.975 & 0.814 & 0.973\\ 
\textbf{SMOTE+VAE} & 0.9919 & 0.961 & 0.795 & 0.975 & 0.815 & \textbf{0.982}\\ 
SMOTE-NC & \multicolumn{1}{c}{-} & \multicolumn{1}{c}{-} & \multicolumn{1}{c}{-} & 0.975 & \multicolumn{1}{c}{-} & \multicolumn{1}{c}{-}\\ 
TGAN & 0.9913 & 0.959 & 0.772 & 0.975 & 0.818 & 0.98\\ 
\end{tabular}

\begin{tabular}{lcccccc} 
 & \textbf{thyroid\_sick} & \textbf{us\_crime} & \textbf{webpage} & \textbf{wine\_quality} & \textbf{yeast\_me2} & \textbf{yeast\_ml8}\\ \toprule 
No OS & 0.9917 & \textbf{0.9242} & 0.969 & 0.847 & 0.897 & 0.593\\ 
CTGAN & 0.9965 & 0.92415 & 0.969 & 0.848 & 0.904 & \textbf{0.621}\\ 
Poly & 0.9965 & 0.92023 & 0.978 & 0.847 & 0.904 & 0.6\\ 
\textbf{Poly+AE} & 0.9956 & 0.92159 & 0.98 & 0.849 & 0.905 & 0.607\\ 
\textbf{Poly+VAE} & \textbf{0.9971} & 0.92071 & 0.976 & \textbf{0.855} & 0.906 & 0.618\\ 
ROS & 0.9965 & 0.92051 & 0.979 & 0.854 & \textbf{0.913} & 0.595\\ 
SMOTE & 0.9966 & 0.9213 & 0.973 & 0.848 & 0.902 & 0.612\\ 
\textbf{SMOTE+AE} & 0.9958 & 0.91994 & \textbf{0.981} & 0.854 & 0.904 & 0.609\\ 
\textbf{SMOTE+VAE} & 0.9968 & 0.92049 & 0.977 & 0.854 & 0.901 & 0.6\\ 
SMOTE-NC & 0.9969 & \multicolumn{1}{c}{-} & \multicolumn{1}{c}{-} & \multicolumn{1}{c}{-} & \multicolumn{1}{c}{-} & \multicolumn{1}{c}{-}\\ 
TGAN & 0.995 & 0.92172 & 0.963 & 0.835 & 0.903 & 0.598\\ 
\end{tabular}
    \caption{Comparison of oversampling methods on $27$ imbalanced datasets. The methods of our framework and the best scores are in bold.}
    \label{tab:results_table}
\end{table*}

In this section we carry out experiments to compare the effectiveness of different oversamplers in addressing imbalanced binary classification challenges. To summarize, it is done by using the various oversamplers to augment the training fold before training the classifier and comparing the classification quality on the test fold.

\textbf{Data} We evaluated the oversampling methods on the $27$ public datasets included in \emph{imbalanced-learn}\footnote{\href{https://imbalanced-learn.org/stable/}{https://imbalanced-learn.org/stable/}}\cite{datasets}, with a varying number of samples, dimensions, and ratios of continuous/discrete features. The preprocessing comprised of normalizing the continuous features to have a unit variance and encoding the discrete features as ordinal integers. The preprocessing was done using \emph{RobustOrdinalEncoder} and \emph{RobustStandardScaler} from \emph{sagemaker-scikit-learn-extension}\footnote{\href{https://github.com/aws/sagemaker-scikit-learn-extension/}{https://github.com/aws/sagemaker-scikit-learn-extension/}}.

\textbf{Metric} ROC-AUC (area under the receiver operating characteristic curve) was used due to it's popularity for binary classification tasks.

\textbf{Oversamplers} We experiment with the oversamplers listed in Section \ref{sec:syntethic_evaluation} with a few modifications: As \emph{Poly} and \emph{SMOTE} do not support discrete features, we applied the common method of encoding the discrete features as ordinal integers and considered them to be continuous, see e.g. \cite{KOVACS2019105662}. In that case, these features were marked as continuous for the classifier. As \emph{SMOTE-NC} supports only datasets that have both numeric and discrete features (fails otherwise), we did not run it for all datasets and it was not included in the summary tables. As the VAE and AE based autoencoders oversamplers performed the best in the synthetic data experiments, we experimented with only them. An additional oversampler introduced is the random oversampler (\emph{ROS}) which augments the dataset by adding random duplicates of samples from the existing minority samples.

\textbf{Method} Each dataset was randomly stratified split into training, validation and test folds with ratios 60\%, 20\% and 20\% respectively. To evaluate the oversampling methods, the training fold was oversampled using each of the oversamplers. The oversamplers that require training, were trained on the training fold with early stopping on the validation fold when possible. Catboost\cite{prokhorenkova2018catboost} was used as the classifier due to its popularity and strong performance on multi-modal tabular data. Catboost was trained on the augmented training fold with early stopping on the (not-augmented) validation fold. Finally, ROC-AUC scores for Catboost's predictions on the validation and test folds were calculated. For each triplet of \{dataset, oversampler, HPs\} the experiment was repeated  $7$ times with different random seeds and different data splits and the ROC-AUC scores were averaged over these $7$ runs.

\begin{table*}
    \footnotesize
    \centering
    \begin{minipage}{0.45\textwidth}
    \begin{tabular}{lcc}
 & \textbf{ROC-AUC} & \textbf{vs baseline} \\ \toprule 
\textbf{Poly+VAE} & {0.923149} & {0.009873}\\ 
\textbf{SMOTE+AE} & {0.921254} & {0.007978}\\ 
\textbf{SMOTE+VAE} & {0.920189} & {0.006913}\\ 
\textbf{Poly+AE} & {0.920062} & {0.006786}\\ 
Poly & {0.920055} & {0.006779}\\ 
SMOTE & {0.919713} & {0.006437}\\ 
ROS & {0.917974} & {0.004698}\\ 
CTGAN & {0.917878} & {0.004602}\\ 
No OS & {0.913276} & {0.000000}\\ 
\midrule 
TGAN & {0.912950} & {-0.000326}\\ 
\bottomrule
\end{tabular}
    \begin{center}(a) Best HPs\end{center}
    \end{minipage}
    \begin{minipage}{0.45\textwidth}
    \begin{tabular}{lcc}
 & \textbf{ROC-AUC} & \textbf{vs baseline} \\ \toprule 
\textbf{Poly+VAE} & {0.916506} & {0.003230}\\ 
\textbf{Poly+AE} & {0.915963} & {0.002687}\\ 
\textbf{SMOTE+VAE} & {0.913652} & {0.000376}\\ 
ROS & {0.913593} & {0.000317}\\ 
No OS & {0.913276} & {0.000000}\\ 
\midrule 
\textbf{SMOTE+AE} & {0.913224} & {-0.000052}\\ 
Poly & {0.912299} & {-0.000977}\\ 
CTGAN & {0.911949} & {-0.001327}\\ 
SMOTE & {0.908617} & {-0.004659}\\ 
TGAN & {0.902926} & {-0.007349}\\ 
\bottomrule
\end{tabular}
    \begin{center}(b) Best validation HPs\end{center}
    \end{minipage}
    \caption{ROC-AUC averaged over all $27$ datasets. In (a) the best oversampler HPs, for each dataset, were used. In (b) the oversampler HPs that yielded the best score on the validation fold, for each dataset, were used. Methods below the mid-line performed worse than than the baseline which is training on the original training data (\emph{No OS}). The average difference from the baseline is presented in the right column. Our proposed methods are in bold.}
    \label{tab:datasets_summary}
\end{table*}

\textbf{Hyper-parameters (HPs)} For each oversampler, we considered several HP configurations including the number (or ratio) of synthetic samples to generate which was 10\%, 20\% or 30\%. For example, \emph{Poly+VAE} had $24$ HP configurations, CTGAN and TGAN had $18$ each and \emph{SMOTE} had $9$.

The choice of HPs can greatly effect the experiment results. A common practice is to use the set of HPs that provides the best results. This practice is explicitly mentioned in \cite{KOVACS2019105662} and is frequently implicitly employed e.g., \cite{Bellinger,doping}. It is reasonable, for example, when the scientist have some previous experience with similar data and knows a-priori how to properly set the HPs. ROC-AUC scores for all oversamplers on all datasets utilizing this practice are displayed in Table \ref{tab:results_table} and aggregation over all datasets is displayed in Table \ref{tab:datasets_summary}a. When no such prior knowledge exist, it is common to select for each dataset and oversampler the HP configuration that maximizes ROC-AUC on the validation fold (and use it to calculate ROC-AUC scores on the test fold). Aggregated results with HPs selected in that manner are presented in Table \ref{tab:datasets_summary}b.

When using the best HPs (Table \ref{tab:datasets_summary}a), the top four oversampler are the VAE and AE based methods. They are followed by the simple \emph{Poly}, \emph{SMOTE} and the random oversampler (\emph{ROS}). \emph{CTGAN} performs slightly worse than \emph{ROS} but still better than training without any oversampling (\emph{No OS}). Finally, oversampling using \emph{TGAN} yields worse results than \emph{No OS}. When selecting HPs based on the validation fold scores (Table \ref{tab:datasets_summary}b), the top performing oversamplers are \emph{Poly+AE} and the two VAE based oversamplers. Other than that, only \emph{ROS} provides better prediction quality than the baseline (\emph{No OS}).

\section{Discussion}
\label{sec:discussion}
Comparison of Tables \ref{tab:datasets_summary}a and \ref{tab:datasets_summary}b reveals that oversampler HPs have a major effect on prediction quality. With best HPs, all but a single oversampler yield a better prediction quality than the baseline (\emph{No OS}) while with best validation HPs, \emph{No OS} was ranked 5\textsuperscript{th} among 10. Moreover, in Table \ref{tab:datasets_summary}a the best model improves ROC-AUC by $0.01$ compared to \emph{No OS} while in Table \ref{tab:datasets_summary}b, the improvement of the best model was only $0.003$.

While the ROC-AUC improvements might seem minor, the immense body of work on oversampling tells a different story. \emph{SMOTE} is widely used and all major random forest algorithms (Catboost\cite{prokhorenkova2018catboost}, lightGBM\cite{lgbm} and XGBoost\cite{XGBoost}) innately support upweighting of the minority class (upweighting is equivalent to \emph{ROS}) thus implying that the small ROC-AUC benefits they provide are important. Over all scenarios, the ROC-AUC improvements of our best performing oversampler \emph{Poly+VAE} were 2-10 times larger than the benefits of \emph{SMOTE} and \emph{ROS}. Therefore, our improvement over these methods is significant.

The autoencoder based approaches using VAE and AE outperformed all other methods in both scenarios with the only exception being \emph{ROS} slightly outperforming \emph{SMOTE+AE} in the best validation HPs scenario. So the balance between \emph{cover} and \emph{error} these methods provide, demonstrated in the artificial datasets study, improves prediction quality for real datasets. Also note that the wrapped oversamplers outperformed their underlying method in all cases. Specifically, \emph{SMOTE+AE} and \emph{SMOTE+VAE} outperform \emph{SMOTE} and \emph{Poly+AE} and \emph{Poly+VAE} outperform \emph{Poly}.

Other than our autoencoder based approaches, the only method that consistently outperformed the baseline is the simple random oversampler (\emph{ROS}). While not improving \emph{cover}, \emph{ROS} provides better prediction quality by increasing the relative weight of the minority samples in the training set. Surprisingly, the relatively simple \emph{SMOTE} and \emph{Poly} did not improve over the baseline without optimized HPs. Note that our experiments agree with the results of \cite{KOVACS2019105662} in that \emph{Poly} demonstrates better performance than \emph{SMOTE} using both HP selection practices.

The advanced GAN based methods recently proposed, CTGAN and TGAN, performed poorly even when compared to the simple interpolation methods. In the artificial data experiments these methods yielded a good \emph{cover} with very high \emph{error} which implies that they tend to generate many synthetic minority points in feature space regions that "belong" to the majority class. Recall that these are general generative methods which are not intended for oversampling the minority and that the label was used as a discrete feature during training. Perhaps better results can be achieved by emphasising the label during model training. Moreover, due to the complexity of these methods, training them requires considerably more hardware and is more expensive compared to our simpler autoencoder based method. In particular TGAN is the slowest and most resource hungry amongst all tested methods, due to the large set of parameters contained within each head representing different columns. TGAN failed training on \emph{isolet} dataset, probably because it includes 617 features - the most in our dataset collection.

\section{Conclusion}
Little attention has been devoted towards generation of synthetic multi-modal tabular data including both continuous and discrete features. We addressed the multi-modal data challenge by proposing a family of oversamplers based on the principle of encoding the data in a dense continuous latent space, interpolate there and map the samples back to the original feature space. We experimented with real and artificial datasets using several autoencoder schemes, two underlying interpolation methods and several other previously proposed oversamplers. Our approach yields a superior prediction quality for real-world datasets and generates better synthetic data compared to all other methods.

We are interested in further exploring the proposed framework in several directions: As we currently train our autoencoders in an unsupervised fashion, we plan to study how the labels can be incorporated in the autoencoder training phase to encourage better interpolation. We also plan to investigate how can our framework be used to benefit other problem types e.g., "balanced" classification or regression. Additionally, modern tabular datasets might include data types such as free text, images, and audio. Support for synthesising of multi-modal data that includes these data types is an interesting avenue for future work.

\subsection*{Acknowledgement}
We would like to thank the authors for allowing the use of datasets provided in \cite{UCI_datasets,coil_dataset,crime_dataset,mammograph_dataset}

\bibliographystyle{abbrv}
\bibliography{references}

\end{document}